\documentclass[]{interact}
\usepackage[markup=underlined]{changes}

\definechangesauthor[color=BrickRed]{JLM}
\definechangesauthor[color=NavyBlue]{JH}

\usepackage{todonotes}
\setcommentmarkup{\todo[color={authorcolor!20},size=\scriptsize]{#3: #1}}

\usepackage{epstopdf}
\usepackage{float}
\usepackage{subfigure} 
\usepackage{subcaption} 

\usepackage{natbib}
\bibpunct[, ]{(}{)}{;}{a}{}{,}
\renewcommand\bibfont{\fontsize{10}{12}\selectfont}

\theoremstyle{plain}

\theoremstyle{definition}

\theoremstyle{remark}

\usepackage{amsmath, amssymb, mathtools}

\begin{document}


\title{Deep Reinforcement Learning for Day-to-day Dynamic Tolling in Tradable Credit Schemes}

\author{
\name{Xiaoyi Wu*\thanks{CONTACT Xiaoyi Wu. Email: xiawu@dtu.dk}, Ravi Seshadri, Filipe Rodrigues, and Carlos Lima Azevedo}
\affil{Technical University of Denmark, Lyngby, DK}
}

\maketitle

\begin{abstract}
Tradable credit schemes (TCS) are an increasingly studied alternative to congestion pricing, given their revenue neutrality and ability to address issues of equity through the initial credit allocation. Modeling TCS to aid future design and implementation is associated with challenges involving user and market behaviors, demand-supply dynamics, and control mechanisms. In this paper, we focus on the latter and address the day-to-day dynamic tolling problem under TCS, which is formulated as a discrete-time Markov Decision Process and solved using reinforcement learning (RL) algorithms. Our results indicate that RL algorithms achieve travel times and social welfare comparable to the Bayesian optimization benchmark, with generalization across varying capacities and demand levels. We further assess the robustness of RL under different hyperparameters and apply regularization techniques to mitigate action oscillation, which generates practical tolling strategies that are transferable under day-to-day demand and supply variability. Finally, we discuss potential challenges—such as scaling to large networks—and show how transfer learning can be leveraged to improve computational efficiency and facilitate the practical deployment of RL‐based TCS solutions.

\end{abstract}


\section{Introduction}

Traffic congestion in urban areas is an important issue that affects environmental sustainability, economic productivity, and social welfare. Congestion pricing (CP) has been widely studied as a means of mitigating these negative externalities and effectively managing demand by discouraging excessive road use. However, the use of CP is often limited by political and social resistance, with critics arguing that it functions as a flat tax that disproportionately impacts low-income travelers and restricts equitable access to transportation.

Given these challenges, Tradable Credit Schemes (TCS) have emerged as a promising alternative. Drawing inspiration from environmental credit trading schemes, TCS allocates a limited number of driving credits or tokens to travelers, who must spend tokens to access roads. Unlike CP, TCS allows travelers to buy and sell credits in a market system, achieving revenue neutrality and potentially enhancing public acceptability through an equitable distribution of initial credits. Recent research has shown that TCS can effectively regulate road usage while addressing inequity \citep{chen2023market,yang2011managing,de2018congestion,seshadri2022congestion}. However, dynamic control mechanisms for adaptive tariff management remained to be studied. 

To bridge this research gap, we propose a day-to-day dynamic tolling framework for TCS in a discrete-time setting, where the regulator sets tolls (in credits) for the next day based on system states observed on previous day(s). Credit market prices adjust from day to day based on the demand and supply relationship. 

Our framework considers a departure-time and mode choice setting, employing a day-to-day version of the classical morning commute problem with a Macroscopic Fundamental Diagram supply model. The dynamic tolling problem is formulated as a Markov Decision Process (MDP) and solved using reinforcement learning (RL) algorithms. We benchmark our approach against scenarios without tolling and strategies optimized through Bayesian optimization (BO) under equilibrium conditions. 

Our results indicate that the RL algorithms achieve similar travel times and social welfare as the BO benchmark, while demonstrating the ability to generalize across different capacities and demand levels. This ability
to learn a tolling policy that is `transferable' is advantageous in traffic management under day-to-day demand and supply variability, opening the door to control
mechanisms suitable for more dynamic conditions. Furthermore, we evaluate algorithm robustness under different hyperparameter settings, which yields insights into how hyperparameters influence RL performance and helps fine-tune the model. Additionally, when applying the RL framework to the dynamic tolling problem in day-to-day operations, we encounter issues of action oscillation in policy training. To address this, we incorporate regularization terms into the policy training, inspired by robotics literature—specifically, the Conditioning for Action Policy Smoothness method proposed by \citet{mysore2021regularizing}.

In summary, the contributions of our work are:

\begin{enumerate}
  \item Formulating the day-to-day dynamic tolling problem for a TCS as an MDP and solving it using a deep RL framework.
  \item Evaluating the proposed framework across different scenarios of demand and supply conditions, assessing its generalization to unseen events.
  \item Assessing algorithm robustness under different hyperparameter settings and policy regularization techniques, shedding light on how these factors influence RL performance. 

\end{enumerate}

The remainder of this paper is organized as follows: Section \ref{section: Literature Review} reviews the relevant literature. Section \ref{section: Methodology} presents a summary of the framework and methodology. Section \ref{section: Experiments} presents the experimental setup, followed by numerical results and discussion in Section \ref{section: Results}. Finally, Section \ref{section: Conclusions} concludes the paper and 
discusses avenues for future research. 

\section{Literature Review} \label{section: Literature Review}

Increasing traffic congestion in urban areas has become a critical concern, with significant impacts on environmental sustainability and social welfare. A widely debated approach to address the negative externalities of road use is CP \citep{pigou1920economics}, recognized for its effectiveness and efficiency. Despite its success in practice \citep{agarwal2015impact, eliasson2021efficient}, CP often encounters political and social resistance, as it is perceived as an unfair tax, potentially hindering road access for low-income travelers and exacerbating mobility inequity \citep{lindsney2001traffic}.

Various approaches have been proposed to mitigate these equity concerns. For example, \citet{liu2009pareto} introduced a
Pareto-improving, revenue-neutral pricing scheme that subsidizes public transit while tolling private car travel on roads. \citet{guo2010pareto} developed a class-anonymous Pareto-improving pricing scheme with revenue-refunding mechanisms in general transportation networks. However, implementing Pareto-improving or revenue-neutral pricing schemes in practice remains highly challenging.

Examining global revenue allocation policies, \citet{carl2016tracking} analyzed data from 40 countries and 16 states, revealing that 70\% of cap-and-trade revenues and 56\% of carbon tax revenues are not redistributed as direct rebates to corporations or individual taxpayers. Instead, these revenues are allocated toward ``green spending'', such as improving energy efficiency and investing in renewable energy, or directed to general government funds. These practical policies suggest that revenue redistribution from tolls or taxes provides limited scope to compensate those adversely affected due to several factors: a portion of the revenue is inevitably consumed by the administrative costs of compensation programs; accurately assessing individual impacts is complex; and self-identified ``losers''
 may exaggerate their losses to maximize compensation \citep{rietveld2003winners, lindsey2020addressing}.
 
 Given the general political resistance to CP, TCS, adapted from environmental management policy \citep{dales1968land}, have been proposed to control road usage in dynamic urban systems \citep{verhoef1997tradeable, goddard1997using, raux2004use}. In TCS, the regulator allocates a set of credits (or tokens) to travelers, who must spend a certain number of tokens before driving \citep{yang2011managing}.
 Tokens can be bought and sold in a market at a price determined endogenously by token demand and supply. Unlike CP or taxation, TCS is inherently revenue-neutral and allows individuals to sell and buy tokens and adjust their travel decisions based on their travel preferences and experiences. \citet{godard2001domestic} summarized four key advantages of tradable token or permit systems for mobility: 1) They are more effective in quantity control without full information on agents’ responses to price/tax. 2) When users are sensitive to quantitative limits, they will plan their trips more carefully within the fixed permit limit. For example, if a driver knows they have a specific number of travel credits, they may be more likely to plan and reduce their trips to stay within this limit. 3) Free endowment of permits enhances public acceptability. 4) Tradability of permits allows traders, other than the government, to benefit by reducing consumption and selling surplus permit quotas. 
 
 Research on TCS has received increased attention from the transportation community, with recent reviews by \citet{provoost2023design} and \citet{servatius2023trading}. Nonetheless, despite the added flexibility introduced by token markets, little attention has been paid to dynamic tariff control mechanisms.
 
Tariff control mechanisms are critical in traffic management. Previous research has studied tolling design in CP contexts based on two main methods: static tolling and dynamic tolling (which can be reactive or adaptive). Static tolling methods are limited in their responsiveness to traffic conditions. Dynamic tolling, while more capable of adjusting toll rates and adapting to the changing urban traffic conditions, requires significant exploration of spatio-temporal features for efficient control \citep{gupta2020real,lentzakis2023predictive}.  \cite{friesz2004dynamic} formulated the dynamic tolling within two-time scales: the within-day and day-to-day scales. Within-day approaches optimize tolls according to actual traffic conditions using equilibrium-based approaches, allowing travelers to adjust their choices in real time. Day-to-day tolling, on the other hand, updates the toll profile daily and focuses on how collective traveling behaviors evolve over time \citep{ wang2015day}.

In the context of day-to-day tolling design, there are two primary approaches \citep{lombardi2021model}: control-based algorithms, which concentrate on traffic flow; and optimization-based algorithms, which aim to maximize specific performance objectives. 

\citet{friesz2004dynamic} pioneered a continuous-time control-based model that calculates optimal time-varying tolls under deterministic dynamics. However, their continuous time formulations are not true ‘day-to-day’ models, and their solutions cannot be used to dynamically price a
network over different days. \citet{rambha2016dynamic} addressed this by formulating a day-to-day pricing mechanism as an MDP in a discrete-time setting, seeking stationary policies that adjust toll rates based on the system’s state. They derived a closed-form solution with explicit transition functions and employed Q-learning in a model-free MDP setup; however, the limited representational capacity of Q-learning restricts its applicability to smaller networks such as the Braess network.

Among optimization-based approaches, meta-heuristics, heuristics, and machine learning methods have been employed for dynamic tolling. For instance, \citet{yang2007steepest} employed heuristic techniques, while \citet{liu2023managing} introduced a surrogate-based optimization framework for TCS tolling. These methods are especially useful when closed-form analytical solutions are impractical.

Several studies have leveraged RL to optimize dynamic tolling. \citet{zhu2015reinforcement} applied an offline RL approach in a lane management context, demonstrating that RL can manage travel demand to minimize total travel time. Although promising, their method did not converge to a stable policy and used discrete action and state spaces, which are less feasible in complex real-world scenarios. \citet{mirzaei2018enhanced} subsequently employed a policy gradient-based RL approach to refine $\Delta$-tolling in real-time, significantly improving toll performance from empirical studies \citep{sharon2017network}.

Recent work on RL-driven dynamic electronic toll collection (DyETC) under CP is relevant to our paper. 
\citet{chen2018dyetc} formulated the real-time tolling optimization process within the ETC systems as an MDP and introduced PG-$\beta$ to constrain the continuous toll rates within practical bounds. Their approach outperformed both Gaussian-based policy gradient methods and fixed or $\Delta$-tolling schemes. To address scalability, \citet{qiu2019dynamic} proposed a cooperative multi-agent RL framework with edge-based graph convolutional networks to capture spatio-temporal correlations, demonstrating scalability and robust performance in realistic settings. Sequently, \citet{wang2022ctrl} adopted a Soft Actor-Critic algorithm with attention-based neural networks to integrate upstream and downstream interactions in a multi-origin, multi-destination network. 

Although these RL frameworks effectively handle real-time demand management in CP, they solely consider single-mode (private vehicle) systems. In contrast, our research expands RL-based dynamic tolling into a TCS that accommodates both private car driving and public transit. This multi-modal perspective demands strategies that consider complex interactions among travelers, transit usage, and credit trading.

In multi-objective contexts, \citet{pandey2020deep} developed an RL approach to balance total travel time and social welfare in express lane management, also testing the transferability of trained models across different data distributions. 

Notably, \citet{sato2022dynamic} conducted a study similar to ours, concentrating on day-to-day dynamic tolling in a morning commute problem. They employed the Deep Deterministic Policy Gradient method to set tolls at each bottleneck in the road network, effectively reducing traffic congestion. However, further examination is needed to assess the approach's robustness, scalability, and generalization, especially in multi-modal transportation systems and under TCS, which our work aims to explore.

Overall, while RL-based approaches show considerable promise for dynamic tolling, existing research has largely focused on single-mode (car driving) CP scenarios. In our work, we extend RL for day-to-day tolling design under the TCS, modeling both car and public transit in the transportation system. We evaluate the RL performance against suitable benchmarks, and examine generalization under unseen conditions and robustness with different hyperparameter settings and regularization techniques.

\section{Methodology}\label{section: Methodology}
\begin{figure}[H]
    \centering   
    \includegraphics[width=5.4 in]
    {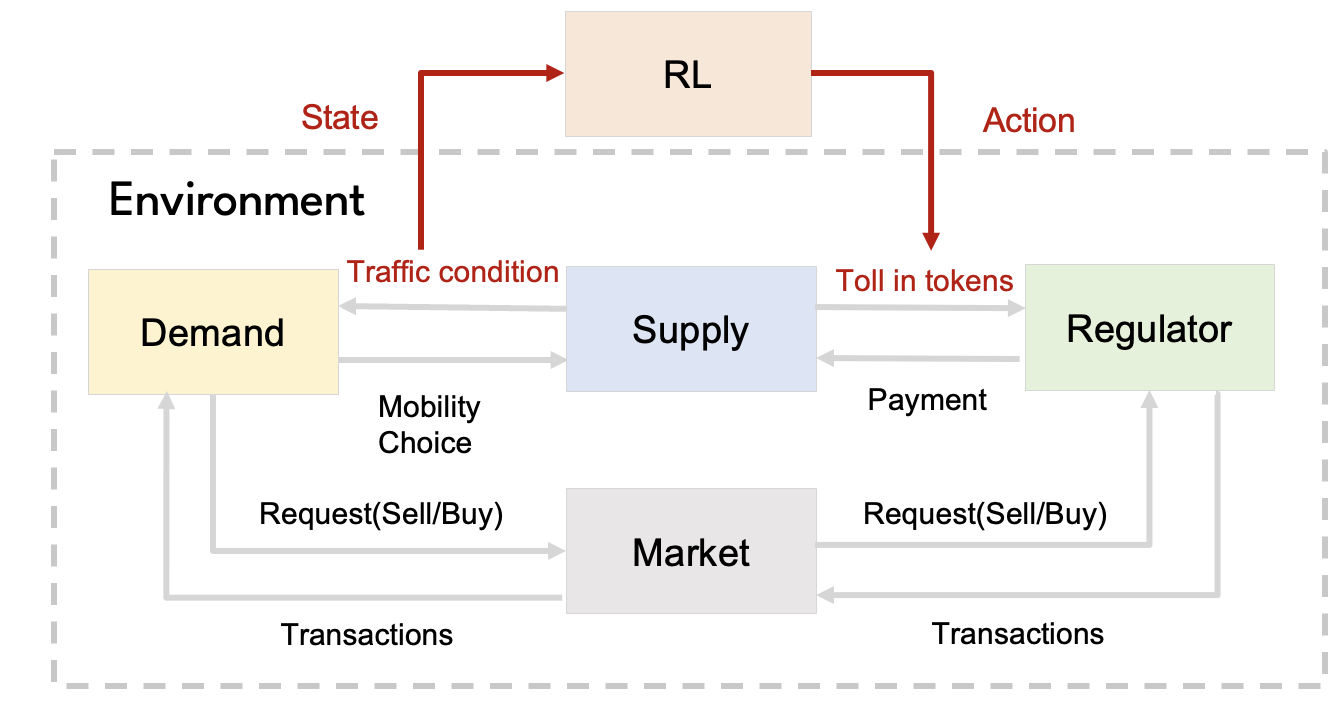}
    \caption{Proposed RL framework}

    \label{fig: Framework}
\end{figure}

We consider a day-to-day dynamic tolling problem for TCS where a regulator adjusts daily time-period specific tolls (in tokens) as a function of the system state (for example, time-dependent departure flows) on the previous day. The daily toll profile in tokens by time of day is parameterized using a Gaussian function with three parameters. This problem is formulated as a finite-horizon MDP and solved using an RL algorithm. 
Broadly, as shown in Figure~\ref{fig: Framework}, the RL agent optimizes its policy (a mapping of the system state to a time-dependent toll profile) based on a reward-action mechanism: the RL agent takes an action (an adjustment to the continuous toll profile over the entire day) and implements it in the environment. 
The environment then simulates the system state for the next day and yields a corresponding reward. The goal of the RL agent is to determine a tolling policy that maximizes the long-term expected reward over a finite time horizon. 

In the remainder of this section, we first describe the environment in more detail in Section \ref{subsec:env}, followed by a formal description of the MDP and the RL algorithm in Section \ref{subsection: Reinforcement Learning}.  

\subsection{Environment}\label{subsec:env}

The environment is a variant of the daily commute problem, where a fixed number of travelers travel from home to work in the morning and return home in the evening. Only the morning commute is explicitly simulated, and the evening trip mirrors the morning as proposed in \citet{chen2023market}. Each traveler has a desired arrival time and an option to choose between public transport (PT) or driving a car, with the flexibility to adjust their departure time if opting to drive. As shown in Figure~\ref{fig: Framework}, the overall environment is adapted from \citet{chen2023market} and \citet{liu2023managing}, and is composed of five modules: 1) \textbf{Demand} simulates travelers' choices between car and PT. If travelers choose to drive, they then select a departure time window. 
2) \textbf{Supply} simulates traffic dynamics based on a trip-based MFD model \citep{liu2023managing}; 3) \textbf{Market} models buying and selling behavior of users under TCS \citep{chen2023market}; 4) \textbf{Regulator} oversees transactions in the token market and adjusts token prices daily in response to the demand and supply variability; 5) \textbf{Day-to-day learning} models travelers' perceived travel costs and expected token account balances through a learning process adapted from \citet{cantarella1995dynamic}. The five modules are explained in turn below. 

\subsubsection{Demand}\label{section: Demand}

The demand model simulates travelers' daily mobility decisions based on a logit-mixture model \citep{ben1985discrete}. Each traveler $n$ has a trip length $TL_n$ and a desired arrival time $\hat{t}_n$. At the beginning of each day, travelers make their mobility decisions, denoted by $ i = \{m, h\} \in \mathcal{I} $, where $ m $ is the transportation mode and $ h $ is the departure time interval. The set of transportation modes $ \mathcal{M}$ includes car ($C$) and public (mass) transit ($PT$). 

Travelers who choose to drive a car must also choose their departure time from a set of intervals $ H_n $, comprising $2\eta$ intervals of size $\Delta h $ centered on the desired departure time $\tilde{t}_{n} $. Consequently, $ H_n $ includes the intervals $ \{\tilde{t}_{n} - \eta \Delta h, \tilde{t}_{n} - (\eta - 1)\Delta h, \ldots, \tilde{t}_{n} + \eta \Delta h\}$. Because the model incorporates income effects, an additional budget constraint applies: a traveler cannot choose any departure interval for which the remaining income (disposable income minus expected cost) is less than zero. 
Thus, we define a set of feasible departure time intervals satisfying the budget constraint $H_n' \subseteq H_n$ under TCS.

For PT users, only one departure time interval $ h_n^{PT} $ is considered (which ensures arrival at the preferred arrival time), assuming constant travel time and average waiting time. Therefore, the overall set of mobility decisions is $ \mathcal{I} = \{C, h|h \in H_n'\} \cup \{PT, h_n^{PT}\} $.

The utility of an individual $n$ choosing mobility decision $i$ is denoted by $U_{in}$, which consists of a systematic utility $V_{in}$ and a random utility component $\epsilon_{in}$. The error term
$\epsilon_{in}$ follows an i.i.d. extreme value distribution with zero mean and individual-specific scale parameter $\mu_n$. Thus, the utility function is given by: 
\begin{equation} 
\begin{aligned}
U_{in}
    & =  V_{in} + \epsilon_{in}. 
\end{aligned}
\end{equation}  
\textbf{Car-driving Utility.} For car drivers,  the utility of departing at $h \in H_n'$  is based on a vector $\bm{\Tilde{\phi}_{in}}$, representing the traveler’s forecast of car travel time, schedule delay costs, and expected costs (considering income effects). The utility function for car drivers is given by:

\begin{equation} 
    \label{eq1}
    \begin{split}
     U_{in} (\bm{\Tilde{\phi}_{in}}) & = V_{in} (\bm{\Tilde{\phi}_{in}}) + \epsilon_{in }  \\
               & = 2\alpha_n \Tilde{\tau}_{in} - \beta_{En} SDE (h, \hat{t}_n, \Tilde{\tau}_{in} ) -\beta_{Ln} SDL (h, \hat{t}_n, \Tilde{\tau}_{in} ) \\
               &\quad + I_n - 2\Tilde{c}_{in} + \lambda_1 \ln (\gamma_1 + I_n -2\Tilde{c}_{in}) + \epsilon_{in},
    \end{split}
\end{equation}  
where:
\begin{itemize}
    \item $\Tilde{\tau}_{in}$ represents individual $n$'s forecast of travel time by car when departing at time window $h$, adjusted through a day-to-day learning process.
    \item $\alpha_n$ is the marginal utility of an additional unit of travel time for individual $n$.
    \item $\beta_{En}$ is the marginal utility of an additional unit of schedule delay early.
    \item $\beta_{Ln}$ is the marginal utility of an additional unit of schedule delay late. The relationship $\beta_{En} \leq \alpha_n \leq \beta_{Ln}$ reflects real-world observations, i.e., arriving late typically incurs more severe consequences \citep{small1982scheduling}. 
    \item $I_n$ represents the individual's disposable income. 
    \item $\gamma_1$ is a constant parameter capturing the nonlinear income effect.
    \item $\lambda_1$ is the co-efficient of the nonlinear income effect adjustment term.
\end{itemize}

$SDE$ represents schedule delay early cost and is calculated as follows:
  \begin{equation}
   SDE (h, \hat{t}_n, \Tilde{\tau}_{in}) = \text{max} (0, (\hat{t}_n - \Delta_{\alpha} - (t_h + \Tilde{\tau}_{in})).
  \end{equation} 
  
$SDL$ represents schedule delay late cost and is calculated as follows:
  \begin{equation}
    SDL (h, \hat{t}_n, \Tilde{\tau}_{in}) = \text{max} (0, (t_h + \Tilde{\tau}_{in}) - \hat{t}_n - \Delta_{\alpha}),
  \end{equation}
where $\Delta_{\alpha}$ represents arrival flexibility: if the traveler arrives outside $[\hat{t}_n - \Delta_{\alpha},\hat{t}_n + \Delta_{\alpha} ]$, they will incur a schedule delay.

The expected cost $\tilde{c}_{in} $ for car driving under the TCS consists of the expected opportunity cost associated with token usage and the fuel cost $c_f $:
  \begin{equation}
    \tilde{c}_{in} = \tilde{R}_{in} + c_f.
  \end{equation}
The opportunity cost $\tilde{R}_{in}$ depends on travelers' forecasted token balance and tariff amount. When a traveler chooses to depart at the time interval $h$ by car, she or he needs to pay the tariff before using the road. Let $t_h$ denote the beginning time of the time window $h$. If the traveler's forecasted token balance $\tilde{x}_n (t_h)$ is higher than the required tariff token amount $T(h)$, then they can sell the extra tokens. Otherwise, they have to buy additional tokens to use the road.
\begin{equation}
\tilde{R}_{in} = 
    \begin{cases} 
    \bigl(Lr- T (h)\bigr) p, & \tilde{x}_n (t_h) \geq T (h) \\ 
    \bigl(T (h) - \tilde{x}_n (t_h)\bigr)p, & \text{otherwise},
    \end{cases}
\end{equation}
\noindent where: 
\begin{itemize}
    \item $p$ is the token price updated based on the demand and supply relationship, which is described in Section \ref{subsubsection: Regulator}.
    \item $r$ is the token allocation rate (tokens are allocated in continuous time, see Section \ref{subsubsection: Regulator}).
    \item $L$ is the token lifetime, i.e., the period after which the token will expire. 
    \item $Lr$ is the full wallet (maximum) amount in the user's token account. 
    \item $T(h)$ is the toll the traveler has to pay to travel at $h$. 

    \item $\tilde{x}_n (t_h)$ is the expected token balance of the traveler, who expects to depart at time $t_h$. 
\end{itemize} 

\textbf{PT Utility.} For PT users, we assume no schedule delay and calculate the utility as:
\begin{equation} 
    \label{eqpt}
    \begin{split}
      U_{in} (\boldsymbol{\Tilde{\phi}_{in}})  & = V_{in} (\boldsymbol{\Tilde{\phi}_{in}}) + \epsilon_{in }  \\
            & = - 2\alpha_n{\tau_{pt}} - 2\beta_{Wn}W_{pt} + I_n -2\Tilde{c}_{in}+ \lambda_1 \ln (\gamma_1+I_n-2\Tilde{c}_{in}) + \epsilon_{in},
    \end{split}
\end{equation} 
where: 
\begin{itemize}
    \item $\tau_{pt}$ is a fixed travel time for PT.
    \item $W_{pt}$ is the expected waiting time.
    \item $\beta_{Wn}$ is the marginal utility of an additional unit of waiting time for PT.  
\end{itemize}

PT users do not need to pay tolls, and hence they can sell all tokens, giving the following expected cost: 
  \begin{equation}
    \Tilde{c}_{in}  = \Tilde{R}_{in} + c_{pt} = -p \cdot L \cdot r + c_{pt}, 
  \end{equation}
  where $c_{pt}$ is the fixed fare cost for taking PT. 

Finally, the probability of individual $n$ choosing alternative $i$ is calculated as follows:
  \begin{equation}
    P_{in} (\boldsymbol{\Tilde{\phi}_{n}})= \frac{\exp \bigl(\mu_n  V_{in} (\boldsymbol{\Tilde{\phi}_{in}})\bigr)}{\sum_{k \in \mathcal{I}_n}\exp \bigl( \mu_n V_{kn} (\boldsymbol{\Tilde{\phi}_{kn}})\bigr)}.
  \end{equation}

\subsubsection{Supply}
The supply network is formulated as a single reservoir with a trip-based Macroscopic Fundamental Diagram (MFD) model. It is adapted from \citet{lamotte2018morning} and described in \citet{liu2023managing}. In the MFD model, traveler $n$'s trip length can be expressed as a function of the average speed on the roads: 
  \begin{equation}
     TL_n = \int_{t_h}^{t_h + \tau_n} v \bigl(n ({t})\bigr) \,dt,
  \end{equation}
  where $t_h$ represents traveler $n$'s departure time, $\tau_n$ is the experienced travel time, $n (t)$ is the accumulation at the current time $t$, and $v\bigl(n ({t})\bigr)$ is a function that describes the dependence between network speed and accumulation.

\subsubsection{Market}
In the TCS system, travelers can buy tokens if their account balance is insufficient and can sell tokens if they choose to depart during off-peak periods or opt for public transit. We assume that travelers can only buy tokens prior to a departure if they are short of tokens and that when they choose to sell tokens, they sell all tokens in their account.  
More details on the implications of these assumptions on the functioning of the market can be found in \citet{chen2023market}. Note that the decision to sell tokens at a given time is closely intertwined with travelers' mobility choices.

\subsubsection{Regulator} \label{subsubsection: Regulator}
Within the TCS, the regulator manages the token market through three primary mechanisms: token allocation, toll rates, and market price adjustment. 

\textbf{Token Allocation.} Tokens are allocated to users at a continuous rate of $r$ to avoid concentrated trading. To prevent speculation and hoarding, each token remains valid for a lifetime $L$, after which it expires. Every user starts with a “full wallet” of $r \times L$ tokens. Once a user’s balance reaches the full wallet threshold of $r \times L$, the oldest token will expire and a new token will be added to the account. 

\textbf{Tolling Profile.}
At the start of each day, the regulator announces a tolling profile that dictates how many tokens drivers must pay based on the time of day. The tolling profile follows a Gaussian-like distribution:
 \begin{equation}
    T(h|M, \mu, \sigma) = M \cdot e^{\frac{-(t_h - \mu)^2}{2\sigma^2}},
 \end{equation}
 where:
 \begin{itemize}
    \item $h$ is the departure time window.
    \item $t_h$ is the beginning time of the time window $h$.
    \item $M$ is the amplitude or peak value of the toll profile.
    \item $\mu$ is the mean or center of the distribution, representing the time of day when the toll is at its peak. 
    \item $\sigma$ is the standard deviation, which controls the spread or width of the toll profile.  
\end{itemize} 

\textbf{Token Price Adjustment.} At the end of each day, the regulator updates the token price $p^{d}$ based on net revenue $K^{d-1}$ from the previous day:
\begin{equation}
p^{d} = \Bigg\{
\begin{array}{ll}
p^{d-1}, & K^{d-1} \in [-\bar{K}, \bar{K}] \\
p^{d-1} - \Delta p, & K^{d-1} < -\bar{K} \\
p^{d-1} + \Delta p, & K^{d-1} > \bar{K}. \\
\end{array}
\end{equation}
Here, $\bar{K}$ is a threshold controlling when price adjustments occur. If the revenue is lower than $-\bar{K}$, the token price will decrease; if the revenue is higher than $\bar{K}$, the price will increase. 
$K^{d-1}$ is the net revenue of buying and selling transactions on day $d-1$. It is calculated as:
\begin{align}
K^{d-1} &= \sum_{n=1}^{N} \biggl\{ \sum_{h \in \{1, \ldots, H \}} \Bigl[ \bigl( T^{d-1}  ( h ) - x_n^{d-1}( t_h ) \bigr)p^{d-1} \mathbb{I}_n^{B, d-1} ( h \mid \boldsymbol{T}^{d-1}) \nonumber\\
&\quad  - x_n^{d-1}( t_h ) p^{d-1} \mathbb{I}_n^{S, d-1} ( h \mid \boldsymbol{T}^{d-1}) \Bigr] \biggr\},  
\label{eq: revenue}
\end{align}
where:
\begin{itemize}
    \item $\boldsymbol{T^{d-1}}$: A vector of discretized tolls on day $d-1$.
    \item $\mathbb{I}_n^{B, d-1}( h \mid \boldsymbol{T^{d-1}})$: is an indicator function showing if traveler $n$ buys extra tokens at time window $h$ due to insufficient balance.
    \item $\mathbb{I}_n^{S, d-1}( h \mid \boldsymbol{T^{d-1}})$: is a function that indicates if traveler $n$ sells tokens at time $h$. 
\end{itemize}

In Equation \ref{eq: revenue}, the first term calculates the revenue obtained by the regulator due to the user buying tokens when the toll $T^{d-1}( h )$ at time interval $h$ is greater than their account balance $x_n^{d-1}( t_h )$ on day $d-1$. The second term calculates the cost to the regulator due to the user selling tokens.

In summary, when the system generates positive net revenue ($K^{d-1}>0$), more tokens are bought than sold, indicating that token demand
exceeds supply, and hence the regulator increases the token price. The threshold $\bar{K}$ is set to ensure that the price adjusts only when the revenue exceeds a certain threshold, thus preventing adjustments for small imbalances in demand and supply. Through the price adjustment scheme, the regulator ensures that total revenue in the
TCS converges to zero at equilibrium, maintaining revenue neutrality.

\subsubsection{Day-to-day learning}

A day-to-day learning process simulates how traveler perceptions evolve across days. We adopt a standard model for updating users' day-to-day perceived travel time based on a weighted sum of historical forecasted travel times $\Tilde{\tau}^{d-1}_n$ and most recent experienced travel times ${\tau}^{d-1}_n$ \citep{cantarella1995dynamic}:
  \begin{equation}
    \Tilde{\tau}^{d}_n = (1-\theta_{\tau}) \Tilde{\tau}^{d-1}_n +\theta_{\tau}{\tau}^{d-1}_n,
    \label{eq: classic d2d}
  \end{equation}
where:
\begin{itemize}
 \item $d$ is the day index.
 \item $\theta_{\tau}$ is a learning weight. 
 \item $\Tilde\tau^{d}_n$ is a vector of traveler $n$'s perceived travel times on day $d$.
 \item $\tau^{d-1}_n$ is a vector of experienced travel times on day $d-1$. 
\end{itemize}

For the chosen departure time, $\tau^{d-1}_n$ is the actual travel time. For all other unchosen departure time intervals within the choice set $H_n$,  we employ the concept of fictional travelers. These fictional travelers are assumed to select these unchosen departure time intervals and calculate travel time without actually affecting accumulation \citep{liu2023managing, lamotte2015dynamic}.

To account for tolling fees' impact on traffic flow expectations (e.g., higher tolls may lower congestion), travelers adapt their travel time perceptions based on the prevailing toll level. We extend the original learning model (Equation \ref{eq: classic d2d}) to incorporate toll-related perceptions as follows:
  \begin{equation}
   \Tilde\tau_n^{d} ({M}) = (1- \theta_t)\Tilde\tau_n^{d-1} ({M}) + \theta_t\tau_n^{d-1} ({M}),
  \end{equation}
where $M$ is the amplitude of the toll profile (discretized in units of 1), and the perception of travel times is updated for the specific toll level in question. For example, if $M=5$ on day $d$, travelers will adjust their perception for the specific toll level of $M=5$, i.e., $\Tilde\tau_n^{d-1} ({M=5})$ and $\tau_n^{d-1} ({M=5})$ whereas perception at other toll levels remains unaffected.

\subsection{Reinforcement Learning}\label{subsection: Reinforcement Learning}

\subsubsection{Markov Decision Process}
The day-to-day dynamic tolling problem in our context is a sequential decision making problem that naturally lends itself to a finite-horizon MDP formulation. An MDP is conventionally defined as a tuple $(\mathcal{S}, \mathcal{A}, \mathcal{P}, \mathcal{R})$ \citep{sutton2018reinforcement}, where $\mathcal{S}$ is the set of states, $\mathcal{A}$ is the set of actions, $\mathcal{P}: \mathcal{S} \times \mathcal{A} \times \mathcal{S} \rightarrow [0,1]$ is the state transition probability distribution, and $\mathcal{R}: \mathcal{S} \times \mathcal{A} \rightarrow \mathbb{R}$ is the reward function. The goal is to find an optimal policy $\pi_{\theta}^{*}:  \mathcal{S} \rightarrow \mathcal{A}$ (a mapping from states to actions), which maximizes the expected cumulative discounted reward over a finite-horizon of length D:
\begin{equation}\label{eq:RL_Obj}
    \pi_{\theta}^{*}(s) = \arg \max_{\pi_{\theta}} \mathbb{E} \left[ \sum_{d=0}^{D} \gamma^d \, \mathcal{R}(s_d, a_d) \,\middle|\, s_0, \pi_{\theta} \right],
\end{equation}
where \(\gamma \in (0,1] \) is the discount factor, and $s_0$ is the initial state.

In our problem, each time step $d$ corresponds to a single day, representing 720 minutes of simulation (clock) time, as the evening commute period is assumed to be a mirror of the morning commute. The time horizon \( D \) is assumed to consist of 60 days, forming one complete episode. The components of the MDP in our framework are defined as follows:
\begin{itemize}   
\item \textbf{Action:} As described in Section \ref{subsubsection: Regulator}, we assume the daily toll for car driving follows a Gaussian-like distribution with a mean $\mu$, a standard deviation $\sigma$ and amplitude value $M$. The action $a_d \in \mathcal{A}$ at time step $d$ is an adjustment to the parameters of the toll profile, i.e., $a_d = (\Delta_{\mu}^{d},\Delta_{\sigma}^{d},\Delta_{M}^{d})$. Thus, the toll profile parameters for day $d+1$ are give by: $\mu^{d+1} = \mu^{d} + \Delta_{\mu}^{d}$, $\sigma^{d+1} = \sigma^{d} + \Delta_{\sigma}^{d}$, and $M^{d+1} = M^{d} + \Delta_{M}^{d}$.  
\item \textbf{State:} The state $s_d \in \mathcal{S} $ at a time step $d$ represents traffic conditions (flows), token price, and toll parameters on day $d$. It is formulated as: 
      \begin{equation}
       s_d = \bigl \langle \boldsymbol{f^d}, p^d , M^d, \mu^d, \sigma^d \bigr \rangle,
      \end{equation}
where $\boldsymbol{f^d}$ is a vector representing the departure flows on day $d$, aggregated over 5-minute intervals, $p^d$ is the token price on day $d$. Observe that the current toll parameter values on the day $d$, $M^d$, $\mu^d$, and $\sigma^d$ are included in the state vector because the actions are formulated as deviations.

\item \textbf{Reward.} 
The reward function $\mathcal{R}(s_d, a_d)$ is formulated to minimize the average individual travel time (AITT) while also encouraging public transit (PT) usage through a term that penalizes deviations of PT usage from capacity. It is formulated as:
 \begin{equation}\label{equation: rw} 
    \mathcal{R}(s_d, a_d) = -\frac{\text{AITT}^d}{\tau_0^{c}} + r_{\text{PT}}^d, 
\end{equation}
where $\tau_0^{c}$ is a constant representing the free-flow travel time by car, and serves as a normalization factor. The term $\text{AITT}^d$ denotes the average individual travel time on day $d$, calculated by:
\begin{equation}
 \text{AITT}^d = \frac{1}{N} \sum_{n=1}^{N} tt_n^d(\boldsymbol{f^d}),
  \end{equation}
where $tt_n^d$ is individual $n$'s travel time on day $d$, and $N$ is the number of travelers. 

The function $r_{\text{PT}}^d$ adjusts the reward based on PT usage to encourage a balanced mode share without exceeding PT capacity limits. It is defined as:
\begin{equation}
r_{\text{PT}}^d = - \left| P_{\text{PT}}^d - P_{\text{PT}} \right|,
\end{equation}
where $P_{\text{PT}}$ is the target PT mode share threshold, set to $0.1$ (or $10\%$) in our experiments, and $P_{\text{PT}}^d$ is the actual PT mode share on day $d$. The term $r_{\text{PT}}^d$ is positive when PT usage is below $10\%$, encouraging more travelers to switch to PT until the desired mode share is achieved. Conversely, it imposes a penalty when PT usage exceeds the threshold, preventing the overloading of the PT system beyond its capacity. As we do not explicitly integrate PT capacity into the environment, this method prevents the underuse or overloading of PT resources. Future work could consider explicitly formulating PT trip costs in the utility function \citep{tang2020modeling}. 
\end{itemize}   

\subsubsection{Proximal Policy Optimization}
Proximal Policy Optimization(PPO) is an on-policy algorithm (within the class of policy-gradient approaches) that operates in two primary phases: the rollout phase and the learning phase \citep{schulman2017proximal}. During the rollout phase, the environment is simulated for a specified number of steps (days in our context), and the resulting trajectories of states are collected. In the learning phase, the policy and value networks are updated based on the collected rollouts. 
The updates are performed to maximize a variant of the objective in Equation \ref{eq:RL_Obj}, which derives from Trust Region Policy Optimization \citep{schulman2015trust}:
\begin{equation}
  L^{CPI}(\theta) = \hat E_t \left[\frac{\pi_{\theta}(a_t | s_t)}{\pi_{\theta_{old}}(a_t|s_t)} \hat A_t \right]  =  \hat E_t \left[ r_t(\theta) \hat A_t \right],
\end{equation}
\begin{equation} \label{CPI}
  \text{subject to } \hat E_t\left[ KL \left[ \pi_{ \theta_{old}}, \pi_{\theta} \right] \right] \leq \delta. 
\end{equation}

Here, $\theta_{old}$ represents the old policy parameters, and $\theta$ are the current policy parameters after the update. $\hat A_t$ denotes the advantage function (refer \cite{schulman2015high} for details), and KL is the KL-divergence between two policies. 
To simplify the above problem, \citet{schulman2017proximal} proposed PPO with two methods based on first-order optimization: 1) Add the KL-divergence as a penalty term in the objective function; 2) Use a clip term in the surrogate objective. We utilize the second method for its simpler implementation.
\begin{equation} \label{CLIP}
    L^{CLIP}(\theta) = \hat E_t \left[(\text{min}(r_t(\theta)\hat A_t, \text{clip}(r_t(\theta), 1-\epsilon, 1+\epsilon) \hat A_t) \right].
\end{equation}
In this equation, when $\hat A_t$ is positive, $\theta$ moves away from $\theta_{old}$ by no more than $(1+\epsilon)\theta_{old}$; otherwise, $\theta$ moves towards $\theta_{old}$ by no less than $(1-\epsilon)\theta_{old}$.

Furthermore, \cite{schulman2017proximal} modify Equation \ref{CLIP} by incorporating an entropy term to encourage exploration:
\begin{equation} \label{CLIP_2}
    L^{CLIP+S}(\theta) = \hat E_t \left[L^{CLIP}(\theta)  + c_1S[\pi_{\theta}](s_t) \right],
\end{equation}
where $c_1$ are coefficients, $S$ is an entropy term. We denote the expression in Equation~\ref{CLIP_2} as $J_{\pi}^\textit{PPO}$ hereafter.

\subsubsection{Action Smoothness}\label{sec:Action_smth}
Action oscillation is a well-known issue when applying deep RL to continuous control tasks \citep{mysore2021regularizing, kobayashi2022l2c2, chen2021addressing, song2023lipsnet}. Controllers in these tasks operate over an infinite action space, which can cause the system to fluctuate or oscillate. Although agents may achieve good cumulative returns during training, action oscillation can be problematic in many real-world applications \citep{chen2021addressing}.

Classical control systems typically address this issue through: 1) reward engineering—modifying the reward manually to induce desired behaviors (\citealt{koch2019flight}; \citealt{carlucho2018adaptive})—however, this typically requires prior knowledge of state and reward information; and 2) filtering policy outputs, i.e., ensuring action smoothness by filtering the RL policy's outputs—a common method in classical control systems \citep{sato2022dynamic}. However, \citet{mysore2021regularizing} found that this approach can change the dynamic response and violate the Markov property, leading to anomalous behavior.

Another branch of methods focuses on reducing the Lipschitz constant of the policy network $\pi_{\theta}$ (a measure of the sensitivity of the network to perturbations in inputs) during training. A smaller Lipschitz constant leads to a smoother loss landscape, thereby reducing action oscillation. \citet{mysore2021regularizing} proposed Conditioning for Action Policy Smoothness (CAPS), which adds regularization to the policy network to achieve smoother actions. \citet{shen2020deep} suggested training the actor network with adversarial perturbations to measure how much the policy output changes in response to small input perturbations. \citet{yu2021taac} introduced a hierarchical structure where one network generates the action distribution and another decides whether to use the generated action. \citet{song2023lipsnet} developed an additional neural network to adjust the Lipschitz constant of the actor network, ensuring policy smoothness. 

In view of these considerations, we apply the CAPS approach to mitigate action oscillation due to its straightforward implementation and simple hyperparameter tuning. Accordingly, we modify the policy optimization objective in the PPO algorithm to:
\begin{equation}
J_{\pi}^{CAPS} = J_{\pi}^{PPO} - \lambda_{T}L_T -  \lambda_{S}L_S .
\end{equation}
Here $ J_{\pi}^{PPO}$ is the surrogate objective function in Equation \ref{CLIP_2} proposed by \citet{schulman2017proximal}, $ \lambda_{T}L_T$ and $ \lambda_{S}L_S$ are two penalties in CAPS to guarantee smoothness in policy network and prevent drastic changes in actions:
\begin{itemize}   
    \item \textbf{Temporal Smoothness}: $L_T$ penalizes the difference between the actions taken in successive states ($s_d$ and $s_{d+1}$), under the assumption that actions taken for consecutive states should be similar. It is defined as $L_T=D(\pi(s_d),\pi(s_{d+1}))$.
    \item \textbf{Spatial Smoothness}: $L_S$ penalizes the difference between the actions taken for similar states ($s_d$ and $\tilde{s}$), where $\tilde{s}$ is a state sampled from the Gaussian distribution $N(s_d, \tilde{\sigma})$, assuming that similar states lead to similar actions.  It is defined as $L_S = D(\pi(s_d), \pi(\tilde{s}))$.
\end{itemize}   

   In these equations, \( D(\cdot, \cdot) \) denotes a distance metric (e.g., Euclidean distance) between two distributions. In our numerical experiments, we investigate how different distance metrics—the L1 norm and the L2 norm—influence the learned policies.

\section{Experiments}\label{section: Experiments}
In this section, we evaluate the performance of the proposed RL framework in two parts. The key inputs and parameters of the environment are summarized in  
Table \ref{table:environment_parameters} and the 
neural network architectures for the policy/value networks and hyperparameters used in PPO are summarized in Table \ref{table:RL_hyperparameters}.

First, we examine the performance of RL with a one-dimensional action space, where the RL agent adjusts only the amplitude ($M$) of the toll profile, using pre-determined values for the mean ($\mu$) and standard deviation ($\sigma$) obtained from BO. We compare our proposed RL framework against BO and NT benchmarks and also evaluate how well RL policies generalize across different scenarios.

Second, we examine the performance of RL with a three-dimensional action space by allowing the RL agent to adjust the amplitude, mean, and standard deviation of the toll profiles simultaneously. Furthermore, we analyze the impact of hyperparameters and smoothness techniques on the RL policies.

To guarantee the availability of at least one feasible time window for each individual that satisfies the budget constraints described in Section \ref{section: Demand}, we set the parameter ranges to $M \in [0, 7]$, $\mu \in [300, 540]$, and $\sigma \in [50, 70]$.

\begin{table}[H]
\centering
\tbl{Environment Parameters}{
\begin{tabular}{p{4cm}p{8cm}p{3cm}}
\toprule
\textbf{Variable} & \textbf{Description} & \textbf{Value} \\
\midrule
$N$ & Population & 7500 \\
$\Delta_t$ & Duration of simulation step (min) & 1 \\
$\Delta_h$ & Duration of departure time step (min) & 1 \\
$\Delta_a$ & Size of desired arrival window (min) & 0 \\
$n(t)$ & Accumulation at time $t$ & - \\
$n_{\text{jam}}$ & MFD capacity (per min) & 7000 \\
$\eta$ & Departure time window size parameter & 60 \\
$v_0^c$ & Free flow speed of car driving (mph) & 45 \\
$v_0^{pt}$ & Free flow speed of public transit (mph) & 18 \\
$\tau_0^c$ & Free flow travel time of car driving (min) & 24 \\
$\tau^{pt}$ & Travel time of public transit (min) & 60 \\
$c_f$ & Fuel cost for driving (\$) & 3.13 \\
$c_{pt}$ & Public transit operation cost (\$) & 2 \\
$v(n(t))$ & Speed function in MFD model & $v_0^c \cdot \left(1 - \frac{n(t)}{n_{\text{jam}}}\right)^2$ \\
$\bar{K}$ & Upper bound of price adjustment & 200 \\
$\Delta p$ & Price change rate (\$/day) & 0.05 \\
$vot$ & Ratio of the value of time to income & 1/4 \\
$dist$ & Travel distance (miles) & 18 \\
$L$ & Token life (min) & 720 \\
$R$ & Allocation rate (token/min) & 0.00269 \\
$FW$ & Full wallet (token) & 1.93680 \\
$\lambda_1$ & Coefficient of nonlinear income effect & 3 \\
$\gamma_1$ & Nonlinear income effect adjustment parameter & 2 \\
$W_{PT}$ & Expected waiting time for public transit (min) & 5 \\
\bottomrule
\end{tabular}}
\tabnote{Parameters are set based on empirical data for the model calibration and may vary depending on the simulation context.}
\label{table:environment_parameters}
\end{table}

\begin{table}[H]
\centering
\tbl{Reinforcement Learning Parameters}{
\begin{tabular}{p{4cm}p{6cm}p{3cm}p{3cm}}
\toprule
\textbf{Parameter} & \textbf{Description} & \textbf{1D Training} & \textbf{3D Training} \\
\midrule
\texttt{n\_steps} & Number of time steps simulated in each environment in the roll-out phase & 60 & 60 \\
\texttt{n\_env} & Number of environments executed in parallel & 10 & 32 \\
Rollout buffer size & Steps collected for a single update, calculated as \texttt{n\_steps} $\times$ \texttt{n\_env} & 2400 & 1920 \\
Batch size & Number of Mini-batches used for gradient update in the learning phase & 600 & 960 \\
\texttt{n\_epoch} & Number of batch shuffles used in gradient update & 10 & 16 \\
\texttt{n\_update} & Number of neural network updates in one learning phase, given by $(\text{Rollout Buffer Size} / \text{Batch Size}) \times \texttt{n\_epoch}$ & 40 & 32 \\
Learning rate & Step size for updating policy parameters & $1 \times 10^{-3}$ & $1 \times 10^{-3}$ \\
\texttt{clip\_range} & Limits the policy update magnitude & 0.2 & 0.2 \\
Discount factor ($\gamma$) & Future reward discounting & 1 & 1 \\
\texttt{ent\_coef} & Controls exploration-exploitation tradeoff & 0.2 & 0.2 \\
\texttt{gae\_lambda} ($\lambda$) & Generalized Advantage Estimation smoothing factor & 1 & 1 \\
Temporal smoothness weight ($\lambda_T$) & Regularization for temporally smooth policy outputs & NA & $1 \times 10^{-4}$ \\
Spatial smoothness weight ($\lambda_S$) & Encourages spatially smooth policy outputs & NA & $1 \times 10^{-4}$ \\
Limit on KL divergence & Threshold for policy divergence control & 0.05 & 0.05 \\
Initial log stdev. (policy network) & Initial standard deviation for policy sampling & -1 & -1 \\
Policy network & Neural network for policy function estimation, shared with Critic Network & \{Tanh(Linear[146,8]), Tanh(Linear[8,8]), Linear[8,1]\} & \{Tanh(Linear[146,8]), Tanh(Linear[8,8]), Linear[8,3]\} \\
Critic network & Neural network for value function estimation, shared with Policy Network & \{Tanh(Linear[146,8]), Tanh(Linear[8,8]), Linear[8,1]\} & \{Tanh(Linear[146,8]), Tanh(Linear[8,8]), Linear[8,1]\} \\
\bottomrule
\end{tabular}}
\label{table:RL_hyperparameters}
\end{table}

\subsection{One-dimensional Action Space}
In this experiment, we train an RL policy with a one-dimensional continuous action space to adjust the amplitude ($M$) of the toll profile. The state vector is defined by $s_d = \bigl \langle \boldsymbol{f^d}, p^d , M^d \bigr \rangle $. 

\subsubsection{Performance Comparison} \label{subsubsection: Performance Comparison}
We train the RL policy using the PPO algorithm with three random seeds, and all the reported metrics are averages across these runs. We compare the RL policy with the following benchmark and baseline at convergence:
\begin{enumerate}
    \item \textbf{No tolling (NT)}: A baseline without tolling.
    \item \textbf{Random}: A baseline with random tolling.
    \item \textbf{Bayesian Optimization (BO)}: We use the BO approach proposed in \citet{liu2021bayesian} to optimize the amplitude of the toll profile. The day-to-day model is simulated for a 60 day period with a constant toll amplitude and the optimization objective is to minimize the average value in Equation \ref{equation: rw} over the last six days of the 60-day simulation. Unlike RL, which optimizes cumulative rewards over the 60-day episode, BO focuses only on maximizing the reward at the equilibrium in the last six days. This method involves 100 iterations with an initial phase that samples 10 data points. 
\end{enumerate}

The performance metrics include AITT, social welfare, mode shares, and credit price stability over the last six days of each episode (assuming the day-to-day process converges to a stationary distribution) to allow for comparisons against the BO.


\subsubsection{Generalization}
Transferability analysis in RL involves evaluating how well an RL policy trained in one environment can adapt and perform in unseen conditions. This analysis is crucial because it demonstrates the generalization and practicality of RL approaches in real-world transportation systems, where supply and demand conditions are often unpredictable due to factors such as day-to-day variability, accidents, weather conditions, and so on. We compare the performance of transferred policies against learn-from-scratch policies across different capacity and demand scenarios.
\begin{enumerate} 
    \item \textbf{Learn-from-scratch policy}: a policy trained and evaluated on the same simulation scenario.
    \item \textbf{Transferred policy}: a policy trained on the original scenario using the simulation parameters outlined in Table \ref{table:environment_parameters} but evaluated on a different scenario.
\end{enumerate}

This comparison allows us to evaluate the effectiveness and adaptability of RL policies to varying and unforeseen circumstances, providing insights into applying RL in real-world traffic management.

\subsection{Three-dimensional Action Space}
We extend the RL model from a one-dimensional to a three-dimensional action space, enabling the agent to adjust three key parameters simultaneously: the amplitude ($M$), mean ($\mu$), and standard deviation ($\sigma$) of the toll profile. This extended action space provides finer control over the tolling strategy, allowing for added flexibility in fluctuating traffic conditions. The state vector is defined by $s_d = \bigl \langle \boldsymbol{f^d}, p^d , M^d, \mu^d, \sigma^d \bigr \rangle $. 


\subsubsection{Hyperparameter Robustness}
In this experiment, we investigate the influence of two key hyperparameters—batch size and the number of update epochs—on the robustness of the RL policy. By systematically varying these parameters, we aim to identify configurations that enhance stability and performance.

\subsubsection{Regularization Robustness}
Training an RL agent within continuous actions presents challenges such as action oscillations and catastrophic forgetting due to the infinite action space, often resulting in convergence to suboptimal policies. To address these issues, we apply actor network regularization techniques \citep{mysore2021regularizing} to improve policy smoothness. We extend their proposed L2-norm temporal and spatial smoothness in CAPS to the L1-norm and examine its performance. This extension allows us to assess the impact of both L1 and L2 regularization on policy smoothness, with the goal of mitigating oscillations in policies.

\section{Results and Discussion}\label{section: Results}

\subsection{One-dimensional Action Space}
\subsubsection{Performance Comparison}\label{section: Benchmark Comparison}
\begin{figure}[H]
\centering
\includegraphics[width=\textwidth]{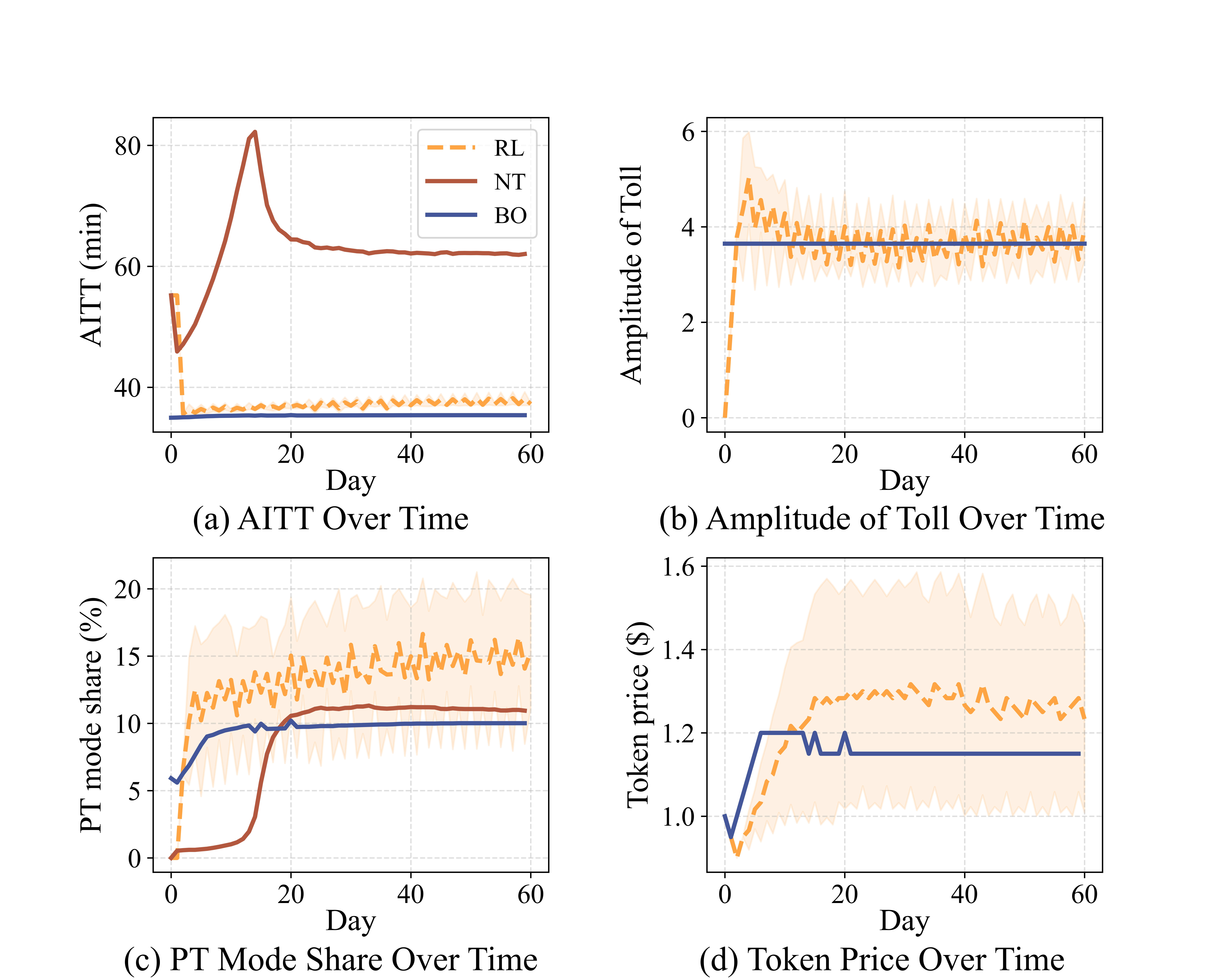}
\caption{\textbf{Benchmark comparison for one dimensional action space.}}
\label{fig: one_dim_benchmark}
\end{figure}

As expected, the results (Table~\ref{table:benchmark_for_1d}) show a significant decrease in AITT in the TCS scenarios compared to the NT scenario, demonstrating effective mitigation of traffic congestion through tolling. The large extent of reductions is a consequence of the congested nature of the network in the baseline scenario. As shown in Figure~\ref{fig: one_dim_benchmark}, BO converges to an optimal toll profile amplitude of 3.65, resulting in an AITT of 35.32 minutes. The toll in the proposed RL approach fluctuates between 3.37 to 3.80 with an average value of 3.66 over the last six days, leading to a corresponding average last-six-day AITT of 37.53 minutes. 

Recall that although we use BO as a benchmark, it solves a fundamentally different problem, namely to determine a toll profile (that is fixed from day to day) that optimizes the objective at equilibrium. In contrast, the RL solves a sequential decision problem where it determines a policy (mapping from states to actions) that maximizes the sum of discounted rewards over the entire episode.   

\begin{table}[H]
\tbl{Model Performance Comparison.}
{\begin{tabular}{lcccc} \toprule
\textbf{Metrics} & \textbf{NT} & \textbf{BO} & \textbf{RL}(1D Action) & \textbf{RL}(3D Action with $T_{L1}$)\\ \midrule
Average AITT (min) & 62.03 & 35.38 & 37.43  & 36.14 \\
Average Car-only AITT (min) & 62.28 & 32.64 & 33.30 & 32.49  \\
Average token price (\$) & NA & 1.15 & 1.30 & 1.33 \\
Average PT mode (\%) & 11.0 & 10 & 15.2  & 13.27  \\
Average social welfare per capita (\$) & NA & 14.59 & 13.87 & 14.58 \\
Average amplitude in toll profile & NA & 3.65 & 3.66 & 3.34 \\
Average mean in toll profile & NA & 443.05 & 443.05 (Given by BO) & 442.37 \\
Average std in toll profile & NA & 63.18 & 63.18 (Given by BO)& 63.33\\

\bottomrule
\end{tabular}}
\label{table:benchmark_for_1d}
\tabnote{Values are average on the last six days within three seeds to represent stable results.}
\end{table}
Figure~\ref{fig: one_dim_benchmark} reveals stark differences in traffic dynamics between the different scenarios. In the NT scenario, severe congestion can be observed during the initial 14 days, due to the absence of tolling measures. Over time, however, the day-to-day learning mechanism prompts travelers to gradually shift toward increased public transit usage and adjusted departure times, effectively mitigating long travel times and schedule delays. In the BO scenario, modest increases in PT mode share and token prices are observed during the first 15 days, which reflect the system's adaptive response to tolling before stabilizing into a steady-state equilibrium.

In the RL framework, we observe three distinct phases in the nature of the tolls over 60 days in the learned policy:
\begin{enumerate}
    \item \textbf{Initial Stage: } Increasing tolls to reduce congestion.
    \item \textbf{Adjustment Stage:} Decreasing tolls as travelers adjust to off-peak departures and transit.
    \item \textbf{Oscillation Stage:} Oscillating tolls periodically. In the last 6 days, the amplitude in the toll profile oscillates between 3.32 and 3.8, averaging 3.66; the token price in turn also fluctuates between 1.27\$ to 1.32\$ with an average value of 1.30\$; and the PT user number oscillates between 14.5\% to 15.32\%, with an average value of 15.22\%.
\end{enumerate}

\subsubsection{Generalization Under Different Capacity}

In addition to the benchmark comparisons, we evaluate the transferability of the proposed RL policies in scenarios with 90\% and 110\% capacity of the baseline capacity. 

First, when the road capacity is reduced to 90\% (Figure 3a), as expected, the NT equilibrium has higher travel times at convergence. 
The transferred RL policy (trained on the baseline capacity scenario) performs reasonably well and converges to a marginally lower toll value than the policy learned from scratch on the 90\% capacity scenario. This suggests that the RL policy is to some degree robust to reductions in capacity.  However, observe that the transferred policy yields actions that oscillate more than the policy learned from scratch. 


When road capacity increases to 110\% (Figure 3b), the opposite trend is observed, i.e., the transferred policy converges to a higher toll than the policy learned from scratch.
The lower tolls in the learn-from-scratch policy result in fewer travelers switching to PT and thus higher congestion levels, and surprisingly, to a lower reward than the transferred policy. This again suggests that the transferred policy is robust to increases in capacity. 
However, the fact that the learn-from-scratch policy does not perform as well as the transferred policy is counterintuitive and could indicate that more hyperparameter tuning is required.


\subsubsection{Generalization Under Different Demand}
We also conduct a transferability analysis under different demand scenarios to evaluate policy robustness under demand fluctuations.

First, when demand decreases to 90\% of the baseline level, as shown in Figure 3c, a drop in travel time is observed in the new NT equilibrium compared to the original NT equilibrium. As for the RL policies under TCS, the learn-from-scratch policy is trained and evaluated in the 90\% demand scenario, while the transferred policy is trained with the original demand and evaluated in the 90\% demand scenario. The results under reduced demand are similar in nature to that of increased capacity. Specifically, the learn-from-scratch policy converges to a lower toll amplitude than the transferred policy, which encourages more people to use the road and results in higher travel times for car drivers, and a lower reward. This finding is again counterintuitive and could indicate that more hyperparameter tuning is required. Nevertheless, it does demonstrate that the trained RL policies are robust to reductions in demand from the baseline. 


In contrast, when demand increases to 110\% of the baseline (Figure 3d), travel times in the NT equilibrium rise accordingly. Here, the learn-from‐scratch policy quickly adopts a higher toll amplitude, effectively curbing road usage and mitigating congestion in the early stages. As expected, it surpasses the transferred policy, which however, yields a comparable travel time and PT usage, thus indicating it is also robust to higher demand levels.  
These results highlight how demand and capacity levels can shape learning dynamics and emphasize the potential of transfer learning to address both demand and supply variability.

\begin{figure}[H]
\centering
\subfigure[Transferability analysis at 90\% capacity.]{%
\resizebox*{14cm}{!}{\includegraphics{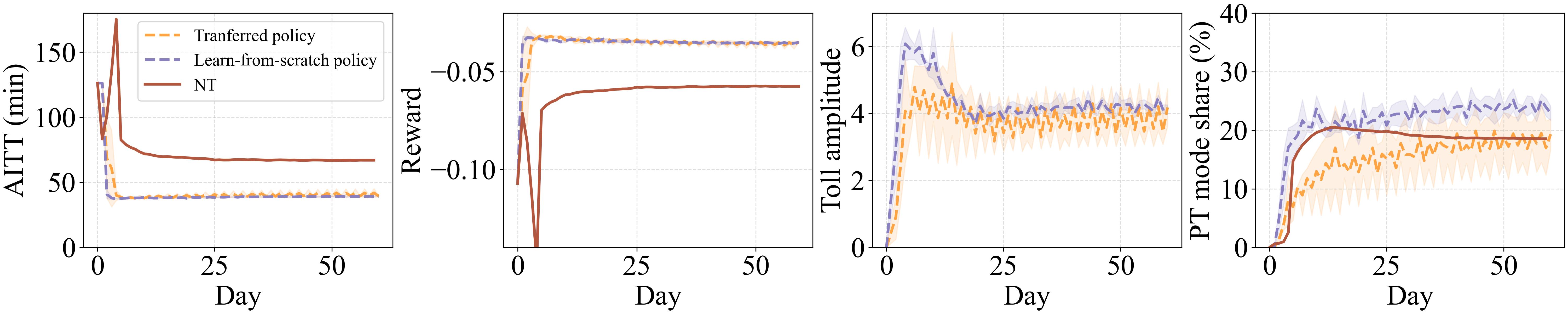}}
\label{fig:transferred_policy_0_9_capacity}
}

\subfigure[Transferability analysis at 110\% capacity.]{%
\resizebox*{14cm}{!}{\includegraphics{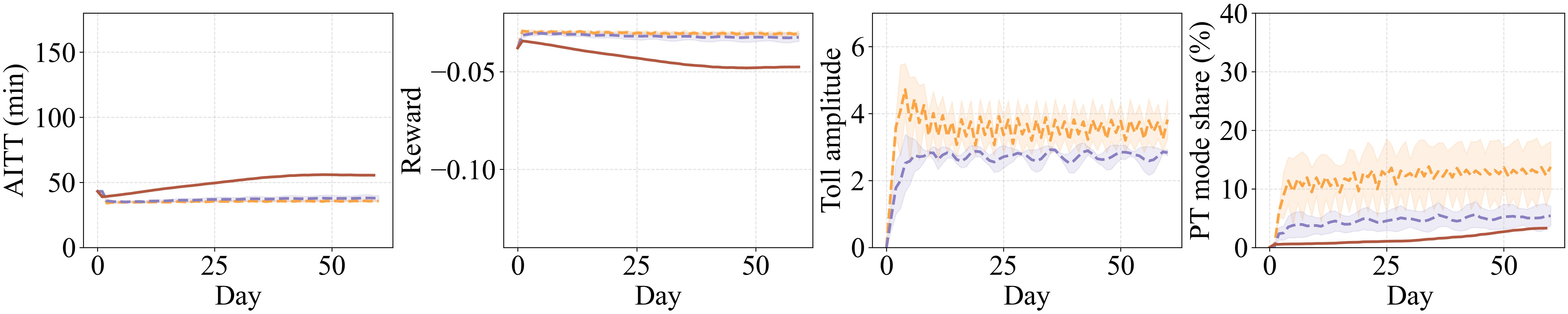}}
\label{fig:transferred_policy_1_1_capacity}}

\subfigure[Transferability analysis at 90\% demand.]{%
\resizebox*{14cm}{!}{\includegraphics{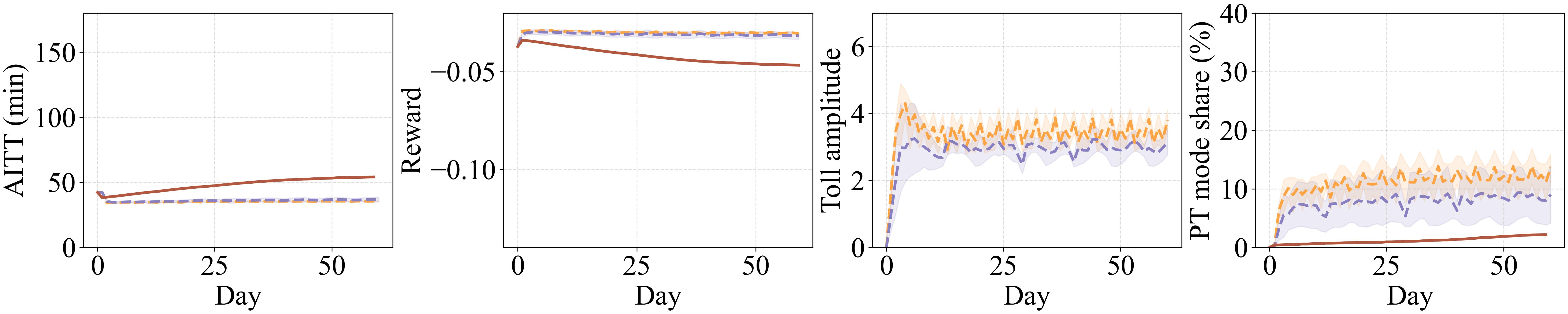}}
\label{fig:transferred_policy_0_9_numOfusers}
}\

\subfigure[Transferability analysis at 110\% demand.]{%
\resizebox*{14cm}{!}
{\includegraphics{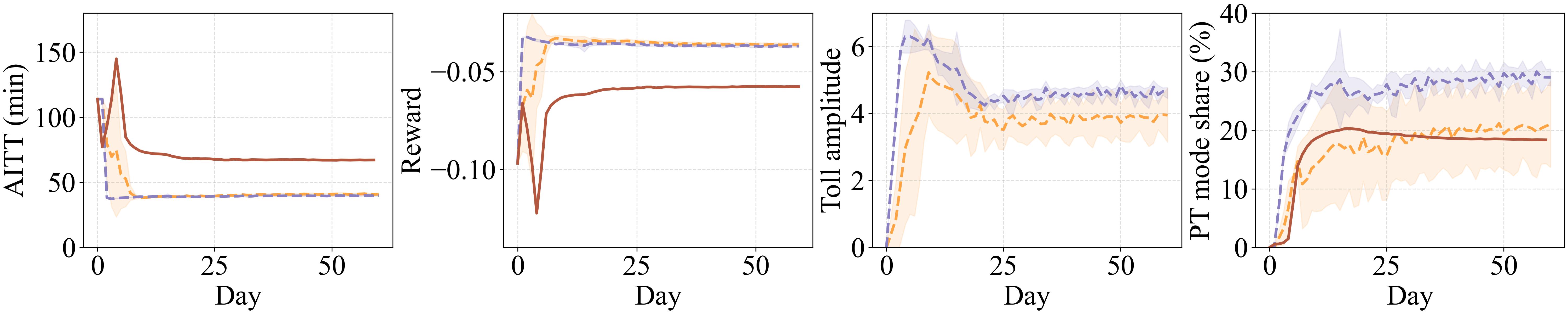}}
\label{fig:transferred_policy_1_1_numOfusers}
}
\caption{\textbf{Comparison of performance between transferred and learn-from-scratch policies under varying levels of congestion and demand.} Both policies exhibit similar performance in terms of AITT and rewards. However, policies trained in highly congested environments adopt more aggressive tolling strategies, with higher peak toll values and faster adjustments in tolls. These strategies result in a greater PT mode share.} 

\label{fig: transferred_policy}
\end{figure}

\subsection{Three-dimensional Action Space}

\subsubsection{Hyperparameter Tuning}
In the following experiments, we first analyze the effect of batch size and number of update epochs on the robustness of the framework.

\begin{figure}[H]
\centering
\includegraphics[width=\textwidth]{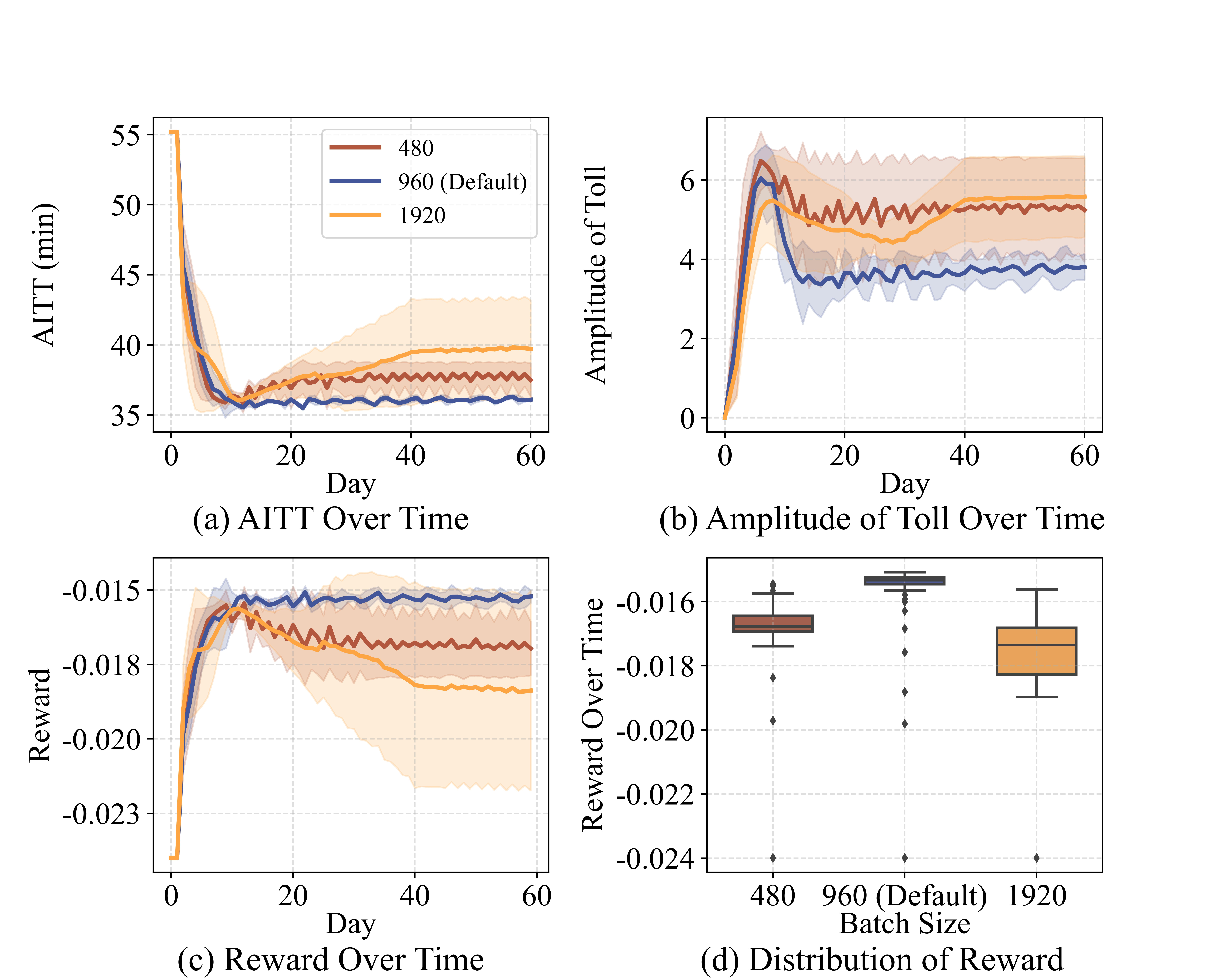}
\caption{\textbf{Impact of Batch Size on PPO Performance.} Overly small batch sizes (e.g., 480) tend to converge quickly to suboptimal, oscillatory policies with highly variable tolling rates, slightly higher AITT, and lower rewards. In contrast, larger batch sizes (e.g., 1920) exhibit greater data diversity, as reflected in large shaded areas for AITT, tolling rates, and rewards. However, they may result in slower convergence and much lower rewards compared to other batch sizes under the same training budget.}%
\label{fig: PPO_batch_size}
\end{figure}

\textbf{Batch Size.} As shown in Figure~\ref{fig: PPO_batch_size}, the performance of the RL policy is extremely sensitive to the PPO algorithm hyperparameters. The original (default) batch size of 960 (one-half of the episode duration) yields the best performance. When the batch size is reduced to 480, PPO converges to a suboptimal oscillatory policy with a high tolling rate, while increasing the batch size to 1920 results in higher variance and slower training convergence speed with a large variance in rewards. This suggests that a small batch size with a higher number of neural network updates per learning phase may lead to suboptimal convergence. Conversely, a large batch size may increase the diversity of data samples, which can be beneficial for exploration in RL, but it also requires longer training time to converge.

\begin{figure}[H]
\centering
\includegraphics[width=\textwidth]{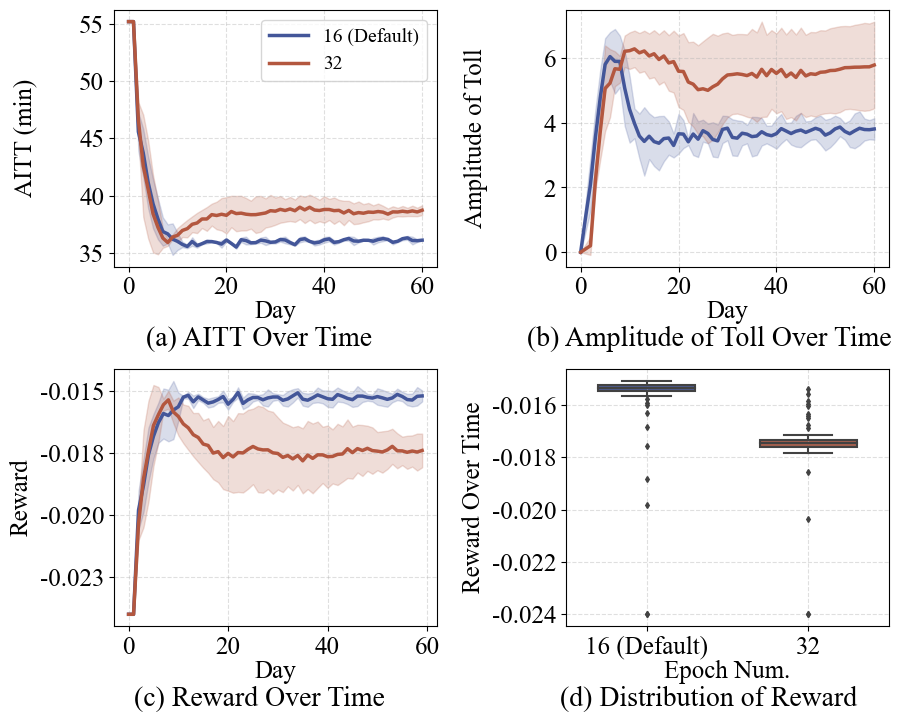}
\caption{\textbf{Impact of Epoch Number on PPO Performance.} In general, increasing the number of epochs allows the RL algorithm to learn more thoroughly from the training data. However, a high epoch number (e.g., 32) combined with a moderate batch size results in oscillatory sub-optimal policies. Balancing the number of epochs and batch size is crucial to ensure stable and robust policy learning.}
\label{fig: PPO_epoch}
\end{figure}

\textbf{Epoch Number.}
We also examine two different epoch sizes—16 and 32—while keeping the batch size fixed at 960, to assess how the number of epochs affects the stability and performance of the RL algorithm. As shown in Figure~\ref{fig: PPO_epoch}, we observe that policies training with a higher epoch number and a moderate batch size yields oscillatory policies with fluctuating amplitude and mean toll values that are clearly sub-optimal. 
In this regard, we observe similar issues with higher epoch numbers as we do with lower batch sizes. This highlights the importance of carefully balancing the number of epochs and batch size to ensure robust learning.

\subsubsection{Regularization}
The numerical results presented above indicate that continuous RL for dynamic day-to-day tolling in multi-modal transportation systems often leads to oscillatory policies: the tolls oscillate significantly in RL trained policies observed in one-dimensional action space and three-dimensional action space training with a small batch size (e.g., 480). This problem with fluctuating tolls causes volatility in the token prices and leads to travelers' switching between driving and PT, and hence, unstable day-to-day dynamics. For example, when tolls are low, more travelers opt to drive, which increases congestion and prompts the RL algorithm to raise tolls. Conversely, when tolls are high, travelers shift to public transit, reducing road congestion and leading the RL algorithm to lower tolls. Increasing the number of epochs with a moderate batch size tends to exacerbate this oscillation. To address these challenges and mitigate the oscillation in actions, we apply actor network regularization techniques \citep{mysore2021regularizing}. We set the batch size to 480 and the number of epochs to 16 (which produced significant oscillations in actions) while keeping the other hyperparameters fixed to the values in Table \ref{table:RL_hyperparameters}.

As noted in Section \ref{sec:Action_smth}, temporal smoothness enforces similar actions across successive days, whereas spatial smoothness enforces similar actions across days with similar states.  

\begin{figure}[H]
\centering
\includegraphics[width=\textwidth]{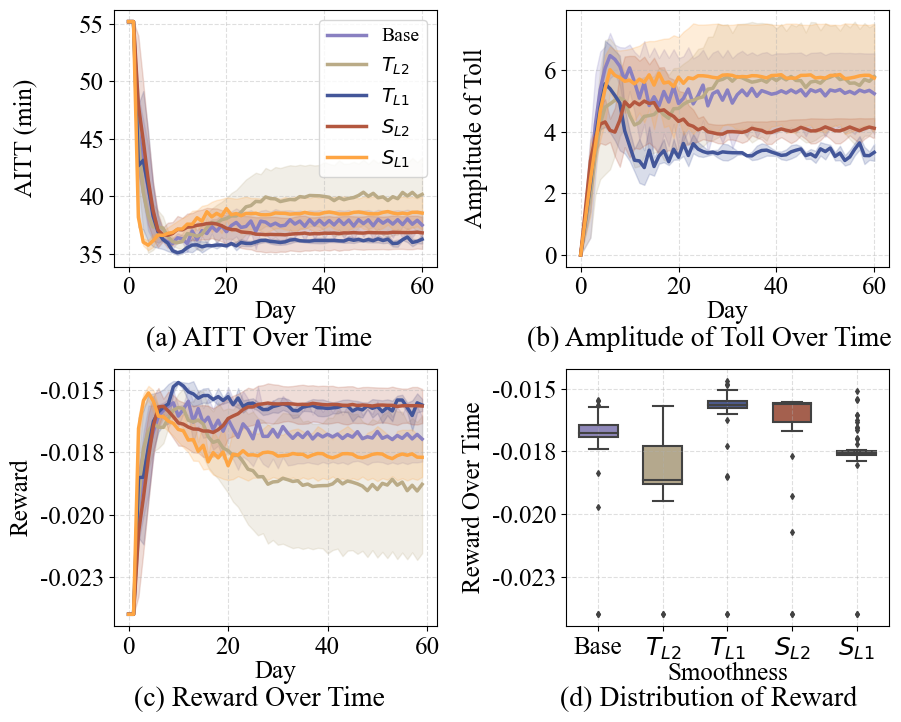}
\caption{ \textbf{Comparison of different regularization techniques}. The graphs show the impact of different regularization techniques—$T_{L1}$ (L1 norm of temporal smoothness), $T_{L2}$
  (L2 norm of temporal smoothness), $S_{L1}$ (L1 norm of spatial smoothness), and $S_{L2}$ (L2 norm of spatial smoothness)—on four metrics over time: AITT, reward, and amplitude of toll. Notably, $T_{L1}$ and $S_{L2}$ yield different policies compared to the RL without regularization, achieving higher rewards, lower AITT, and smoother toll transition.} 
  \label{fig: PPO_reg}
\end{figure}

Figure~\ref{fig: PPO_reg} shows that applying the L1-norm of temporal smoothness and the L2-norm of spatial smoothness in PPO improves both episodic returns and policy smoothness, with a reduction in variance. Thus, these approaches may be useful in real-world applications where action oscillations are undesirable. Furthermore, it can be seen that the different approaches produce tolling policies that are qualitatively different, which also has implications for real-world implementation. Policies with L1-norm of temporal smoothness develop three-stage policies as the original PPO: First, the amplitude increases rapidly to a high value in the initial stage. Then, it decreases as travelers adjust their departure times, ultimately converging to a value of around 3.33. In contrast, policies using the L2-norm for spatial smoothness adopt a different strategy, characterized by smoother transitions and an overall higher tolling rate. The amplitude initially rises to a high level and then gradually decreases, eventually stabilizing within the range around 4.11. Developing an intuition to explain the differences between the four different types of regularization is difficult. It is evident from the results that the four regularization approaches function differently and that their performance may be context-specific. Thus, the choice of regularization should be approached in the same manner as hyper-parameter tuning.  


\section{Conclusions}\label{section: Conclusions}
In this paper, we propose an RL-based framework to optimize day-to-day dynamic tolling under Tradable Credit Schemes. 
Numerical experiments demonstrate that the RL-based approach efficiently mitigates congestion through optimal tolling strategies relative to suitable benchmarks. We assess the generalizability of the RL approach across various scenarios with different road capacities and demand levels. The transferability analysis shows that our RL policies respond well to fluctuations in demand and supply, indicating that the framework generalizes effectively under uncertain traffic conditions and during unusual events.



A key challenge that we observe with the application of RL to day-to-day dynamic systems is action oscillation during the training process, which results in sub-optimal solutions and diminished RL performance. To alleviate this issue, we apply regularization techniques, extending the CAPS from L2 norms to L1 norms. An ablation study further underscores the effectiveness of different smoothness approaches on policy improvement. The findings indicate that different regularization techniques can generate diverse solutions in tolling design adaptable to different management needs.

Future research could examine the application of the framework to environments with more realistic modeling of travel behavior, public transportation, and network topology, thus bridging the simulation-to-real-world gap. Transfer learning and regularization techniques may be helpful in this regard and could improve the practical applicability of the RL-based dynamic tolling frameworks for real-world traffic management.

\section*{Acknowledgements}
This research was conducted as part of the Demand Responsive Electrified Multimodal Transit Systems project, funded by the DTU (Technical University of Denmark)–KTH Royal Institute of Technology Alliance. Additional support was provided through scholarships from the Otto Mønsted Foundation, and the Danish Agency for Higher Education and Science (DAHES) and the Massachusetts Institute of Technology (MIT) partnership.

\bibliographystyle{tfcad}
\bibliography{bibliography}
\end{document}